\documentclass{article}

\usepackage{microtype}
\usepackage{graphicx}
\usepackage{subfigure}

\usepackage{bbm}
\usepackage{url}
\usepackage{amsmath,amssymb}

\usepackage{graphicx}
\usepackage{subfigure} 
\usepackage{wrapfig}

\usepackage{array}
\usepackage{multicol}%
\usepackage{algpseudocode}

\usepackage{algorithm}
\usepackage{booktabs} 

\usepackage{amsthm}
\newtheorem{amsdefi}{Definition}

   \PassOptionsToPackage{numbers, compress}{natbib}

    \usepackage[preprint]{neurips_2019}

\usepackage[utf8]{inputenc} 
\usepackage[T1]{fontenc}    
\usepackage{hyperref}       
\usepackage{color}
\usepackage{url}            
\usepackage{booktabs}       
\usepackage{amsfonts}       
\usepackage{nicefrac}       
\usepackage{microtype}      

\title{A Benders Decomposition Approach to \\Correlation Clustering}

\author{%
  Margret Keuper \\ 
   University of Mannheim\\
   Baden Wurttemberg, Germany \\
  \texttt{keuper@uni-mannheim.de} \\
  \And
  Jovita Lukasik \\
  University of Mannheim \\
  Baden Wurttemberg, Germany \\
  \texttt{jovita@informatik.uni-mannheim.de} \\
  \And
  Maneesh Singh \\
  Verisk \\
  Jersey City, New Jersey, USA \\
  \texttt{maneesh.singh@verisk.com} \\
  \And
  Julian Yarkony \\
  Verisk\\
  Jersey City, New Jersey, USA \\
  \texttt{julian.yarkony@verisk.com} \\
}

\begin{document}

\maketitle

\begin{abstract}
We tackle the problem of graph partitioning for image segmentation using correlation clustering (CC), which we treat as an integer linear program (ILP). We reformulate optimization in the ILP so as to admit efficient optimization via Benders decomposition, a classic technique from operations research. Our Benders decomposition formulation has many subproblems, each associated with a node in the CC instance's graph, which can be solved in parallel. Each Benders subproblem enforces the cycle inequalities corresponding to edges with negative (repulsive) weights attached to its corresponding node in the CC instance. We generate Magnanti-Wong Benders rows in addition to standard Benders rows to accelerate optimization. Our Benders decomposition approach provides a promising new avenue to accelerate optimization for CC, and, in contrast to previous cutting plane approaches, theoretically allows for massive parallelization.
\end{abstract}

\section{Introduction}

Many computer vision tasks involve partitioning (clustering) a set of observations into unique entities. A powerful formulation for such tasks is that of (weighted) correlation clustering (CC). CC is defined on a sparse graph with real valued edge weights, where nodes correspond to observations and weighted edges describe the affinity between pairs of nodes.  

For example, in image segmentation (on superpixel graphs), nodes correspond to superpixels and edges indicate adjacency between superpixels. The weight of the edge between a pair of superpixels relates to the probability, as defined by a classifier, that the two superpixels belong to the same ground truth entity.  This weight is positive, if the probability is greater than $\frac{1}{2}$ and negative if it is less than $\frac{1}{2}$.  The magnitude of the weight is a function of the confidence of the classifier.   

The CC cost function sums up the weights of the edges separating connected components (referred to as entities) in a proposed partitioning of the graph. Optimization in CC partitions the graph into entities so as to minimize the CC cost. CC is appealing, since the optimal number of entities emerges naturally as a function of the edge weights, rather than requiring an additional search over some model order parameter describing the number of clusters (entities) \cite{PlanarCC}. 

Optimization in CC is NP-hard for general graphs \cite{corclustorig}. Previous methods for the optimization of CC problems such as described in \citet{ilpalg,Andres2012} and \citet{nowozin2009solution} are based on linear programming with cutting planes. They do not scale easily to large CC problem instances and are not easily parallelizable. The goal of this paper is to introduce an efficient mechanism for optimization in CC for domains, where massively parallel computation could be employed. 

In this paper we apply the classic Benders decomposition from operations research~\cite{benders1962partitioning} to CC for computer vision. Benders decomposition is commonly applied in operations research to solve mixed integer linear programs (MILP) that have a special but common block structure. Benders decomposition partitions the variables in the MILP between a master problem and a set of subproblems. The block structure requires that no row of the constraint matrix of the MILP contains variables from more than one subproblem. Variables explicitly enforced to be integral lie only in the master problem. 

Optimization in Benders decomposition is achieved using a cutting plane algorithm. Optimization proceeds with the master problem solving optimization over its variables. The subsequent solution of the subproblems can be done in parallel and provides primal/dual solutions over their variables conditioned on the solution to the master problem. The dual solutions to the subproblems provide constraints to the master problem. Optimization continues until no further constraints are added to the master problem.

Benders decomposition is an exact MILP programming solver, but can be intuitively understood as a coordinate descent procedure, iterating between the master problem and the subproblems.  Here, solving the subproblems not only provides a solution for their variables, but also a lower bound in the form of a hyper-plane over the master problem's variables. This lower bound is tight at the current solution to the master problem.   

Benders decomposition is accelerated using the seminal operations research technique of Magnanti-Wong Benders rows~(MWR)~\cite{magnanti1981accelerating}.  MWR are generated by solving the Benders subproblems with an alternative (often random) objective under the hard constraint of optimality (possibly within a factor) regarding the original objective of the subproblem. \color{black} 

Our contribution is the use of Benders decomposition with MWR to tackle optimization in CC. This allows for massive parallelization, in contrast to classic approaches to CC such as in \citet{ilpalg}.
%
%
%
%
\section{Related Work}
\label{Sec_relWork}
Correlation clustering has been successfully applied to multiple problems in computer vision including image segmentation, multi-object tracking, instance segmentation and multi-person pose estimation. The classical work of \citet{ilpalg} models image segmentation as CC, where nodes correspond to superpixels. \citet{ilpalg} optimize CC using an integer linear programming (ILP) branch-and-cut strategy \color{black} which precludes parallel execution\color{black}. \citet{highcc} extend CC to include higher-order cost terms over sets of nodes, which they solve using an approach similar to \cite{ilpalg}. A parallel optimization scheme for complete, unweighted graphs has been proposed by \citet{Pan:2015:PCC:2969239.2969249}. This approach relies on random sampling and only provides optimality bounds.

\citet{PlanarCC} tackle CC in the planar graph structured problems commonly found in computer vision. They introduce a column generation \cite{cuttingstock,barnprice} approach, where the pricing problem corresponds to finding the lowest reduced cost 2-colorable partition of the graph, via a reduction to minimum cost perfect matching \cite{fisher2,maxcutuni,kolmblos}. This approach has been extended to hierarchical image segmentation in \citet{HPlanarCC} and to specific cases of non-planar graphs in \citet{nips15work,Zhang14a,AndresYarkony2013}.  

Large CC problem instances such as defined in \citet{bjoern1,bjoern2} and \citet{fusion_move} are usually addressed by primal feasible heuristics~\cite{CGC,fusionMoves,Kardoost_Keuper_ACCV2018,bjoern1,swoboda-2017}. Such approaches are highly relevant in practice whenever the optimal solution is out of reach, but they do not provide any guarantees on the quality of the solution. \color{black}

\citet{trackerbio} tackles multi-object tracking using a formulation closely related to CC, where nodes correspond to detections of objects and edges are associated with probabilities of co-association.The work of \citet{deepcut2} and \citet{deepcut1} build on \citet{trackerbio} in order to formulate multi-person pose estimation using CC augmented with node labeling.

Our work is derived from the classical work in operations research on Benders decomposition~\cite{benders1962partitioning,birge1985decomposition,geoffrion1974multicommodity}. Specifically, we are inspired by the fixed charge formulations of \citet{cordeau2001benders}, which solves a mixed integer linear program over a set of fixed charge variables (opening links) and a larger set of fractional variables (flows of commodities from facilities to customers in a network) associated with constraints. Benders decomposition reformulates optimization so as to use only the integer variables and converts the fractional variables into constraints.  These constraints are referred to as Benders rows.  Optimization is then tackled using a cutting plane approach. Optimization is accelerated by the use of MWR~\cite{magnanti1981accelerating}, which are more binding than the standard Benders rows.

Benders decomposition has recently been introduced to computer vision (though not for CC), for the purpose of multi-person pose estimation~\cite{wang2017exploiting,wang2018accelerating,yarkony2018accelerating}. In these works, multi-person pose estimation is modeled so as to admit efficient optimization, using column generation and Benders decomposition jointly. The application of Benders decomposition in our paper is distinct regarding the problem domain, the underlying integer program and the structure of the Benders subproblems.
%

\section{Standard Correlation Clustering Formulation}
\label{Sec_origForm}
\label{Sec_origForm}
In this section, we review the standard optimization formulation for CC \cite{ilpalg}, which corresponds to a graph partitioning problem w.r.t. the graph $\mathcal{G} = (\mathcal{V},\mathcal{E})$. This problem is 
defined by the following binary edge labeling problem. 
\begin{amsdefi}
Given a graph $\mathcal{G}= (\mathcal{V}, \mathcal{E})$ with nodes $v \in \mathcal{V}$ and undirected edges $(v_i,v_j) \in \mathcal{E}$. A label $x_{v_i v_j} \in \{0,1\}$ indicates with $x_{v_i v_j}=1$ that the nodes $v_i,v_j$ are in separate components and is zero otherwise. Given the edge weight $\phi_{v_iv_j} \in \mathbb{R},$ the binary edge labeling problem is to find an edge label $\mathbf{x}= (x_{v_i v_j})\in \{0,1\}^{\vert \mathcal{E} \vert}$, for which the total weight of the cut edges is minimized:
\begin{align}
\min_{\substack{\mathbf{x} \in \{0,1\}^{|\mathcal{E}|}   }} & \sum_{(v_i,v_j)\in \mathcal{E}^-} -\phi_{v_iv_j} (1-x_{v_iv_j})
+\sum_{(v_i,v_j)\in \mathcal{E}^+} \phi_{v_iv_j} x_{v_iv_j} \label{origObj} \tag{$\mathrm{CC}_1$}  \\
\textnormal{s.t.} & \sum_{(v_{i},v_j)\in \mathcal{E}^+_c}x_{v_iv_j} \geq x_{v_i^cv_j^c} \quad \forall c\in \mathcal{C}, \label{origConstraint}
\end{align}
where $\mathcal{E^-}, \mathcal{E^+}$ denote the subsets of $\mathcal{E}$, for which the weight  $\phi_{v_iv_j}$ is negative and non-negative, respectively, $\mathcal{C}$ is the set of undirected cycles in $\mathcal{E}$ containing exactly one member of $\mathcal{E^-}$, $(v^c_i, v^c_j)$ is the edge in $\mathcal{E^-}$ associated with cycle $c$ and $\mathcal{E}^+_{c} \subseteq \mathcal{E^+}$ associated with cycle $c$.
\end{amsdefi}
Note that the graph $\mathcal{G}$ defined by $\mathcal{E}$ is very sparse for real problems~\cite{PlanarCC}. Also we refer to an edge $(v_i,v_j)$ with $x_{v_iv_j}=1$ as a \emph{cut} edge.
The objective in Eq.~\eqref{origObj} is to minimize the total weight of the cut edges. The constraints in Eq.~\eqref{origConstraint} ensure that, within every cycle of $\mathcal{G}$, the number of cut edges can not be exactly one. This enforces the labeling $\mathbf{x}$ to decompose $\mathcal{G}$ such that cut edges are exactly those edges that straddle distinct components. We refer to the constraints in Eq. \eqref{origConstraint} as cycle inequalities.

 Solving Eq. \eqref{origObj} is intractable due to the large number of cycle inequalities. \citet{ilpalg} generates solutions by alternating between solving the ILP over a nascent set of constraints $\hat{\mathcal{C}}$ (initialized empty) and adding new constraints from the set of currently violated cycle inequalities. Generating constraints corresponds to iterating over $(v_i,v_j) \in \mathcal{E}^-$ and identifying the shortest path between the nodes $v_i,v_j$ in the graph with edges $\mathcal{E}\setminus (v_i,v_j)$  and weights equal to $\mathbf{x}$.  If the corresponding path has total weight less than $x_{v_iv_j}$, the corresponding constraint is added to $\hat{\mathcal{C}}$.  The LP relaxation of Eq. \eqref{origObj}-\eqref{origConstraint} can be solved instead of the ILP in each iteration until no violated cycle inequalities exist, after which the ILP must be solved in each iteration.

We should note that earlier work in CC for computer vision did not require that cycle inequalities contain exactly one member of $\mathcal{E}^-$, which is on the right hand side of Eq. \eqref{origConstraint}.  It is established with $\textnormal{Lemma} (1)$ in \citet{yarkony2014parallel}, that the addition of cycle inequalities, that contain edges in $\mathcal{E}^-$, $\mathcal{E}^+$ on the left hand side, right hand side of Eq. \eqref{origConstraint}, respectively, do not tighten the ILP in Eq.~\eqref{origObj}-\eqref{origConstraint} or its LP relaxation. 

In this section, we reviewed the baseline approach for solving CC in the computer vision community. In the subsequent sections, we rely on the characterization of CC in Eq. \eqref{origObj}-\eqref{origConstraint}, though not on the specific solver of \citet{ilpalg}.

\section{Benders Decomposition for  Correlation Clustering}
\label{Sec_BendersForm}
In this section, we introduce a novel approach to CC using Benders decomposition (referred to as BDCC).
%
Our proposed decomposition is defined by a minimal vertex cover on $\mathcal{E^-}$ with members $\mathcal{S} \subset \mathcal{V}$ indexed by $v_s$. Each $s \in \mathcal{S}$ is associated with a Benders subproblem and $v_s$ is referred to as the root of that Benders subproblem. Edges in $\mathcal{E}^-$ are partitioned arbitrarily between the subproblems, such that each $(v_i,v_j) \in \mathcal{E}^-$ is associated with either the subproblem with root $v_i$ or the subproblem with root $v_j$. Here, $\mathcal{E}^-_s$ is the subset of $\mathcal{E}^-$ associated with subproblem $s$. The subproblem with root $v_s$ enforces the cycle inequalities $\mathcal{C}_s$, where $\mathcal{C}_s$ is the subset of $\mathcal{C}$ containing edges in $\mathcal{E}^-_s$. We use $\mathcal{E}^+_s$ to denote the subset of $\mathcal{E}^+$ adjacent to $v_s$. 
%
 %
 
 In this section, we assume that we are provided with $\mathcal{S}$, which can be produced greedily or using an LP/ILP solver.  

Below, we rewrite Eq.~\eqref{origObj} using an auxiliary function $Q(\phi,s,\mathbf{x})$.  Here $Q(\phi,s,\mathbf{x})$ provides the cost to alter $\mathbf{x}$ to satisfy all cycle inequalities in $\mathcal{C}_s$, by increasing/decreasing $x_{v_iv_j}$ for $(v_i,v_j)$ in $ \mathcal{E}^+$/$\mathcal{E}^-_s$, respectively. Below we describe the changes of the master's problem edge labeling $\mathbf{x}$, which is based on the edge labeling of each Benders subproblem $\mathbf{x}^s   = (x^s_{v_iv_j}) \in \{0,1\}^{\vert s\vert }$, where $\vert s \vert$ is the number of edges in the subproblem $s$.
\begin{align}
\label{objAug}
\eqref{origObj} \rightsquigarrow  \eqref{objAug}\colon\min_{\mathbf{x} \in \{0,1\}^{\vert \mathcal{E}\vert }}\sum_{(v_i,v_j)\in \mathcal{E}^-} -\phi_{v_iv_j} (1-x_{v_iv_j}) +\sum_{(v_i,v_j) \in \mathcal{E}^+} \phi_{v_iv_j} x_{v_iv_j}+\sum_{s \in \mathcal{S}}Q(\phi,s,\mathbf{x}), \tag{$\mathrm{CC}_2$}
\end{align}
where $Q(\phi,s,\mathbf{x})$ is defined as follows.  
\begin{eqnarray}
\label{QPhiDef}
Q(\phi,s,\mathbf{x})&=& \min_{\substack{\mathbf{x}^s \in \{0,1\}^{|s|}  \color{black} }} \sum_{\substack{(v_i,v_j)\in \mathcal{E}^-_s}} -\phi_{v_iv_j} (1-x^s_{v_iv_j})
+\sum_{\substack{(v_i,v_j)\in \mathcal{E}^+} } \phi_{v_iv_j} x^s_{v_iv_j}  \\
\mbox{  s.t.  }&& \sum_{(v_i,v_j)\in \mathcal{E}^+_c}x_{v_iv_j}+x^s_{v_iv_j} \geq x_{v^c_iv^c_j}-(1-x^s_{v^c_iv^c_j}) \quad \forall c\in \mathcal{C}_s. \nonumber 
\end{eqnarray}

%

We now construct a solution $\mathbf{x}^*=\{x_{v_iv_j}^*, (x_{v_iv_j}^{s*})_{s\in \mathcal{S}}\}$ for which Eq.~\eqref{objAug} is minimized and all cycle inequalities are satisfied. We 
 start from  a given solution $ \mathbf{x}=\{x_{v_iv_j} , (x_{v_iv_j}^s)_{s \in \mathcal{S}}\}$ and proceed as follows. 
\begin{align}
    x_{v_iv_j}^* \overset{\vartriangle}{=}  \min(x_{v_iv_j},x^s_{v_iv_j}) \quad \forall (v_i,v_j) \in \mathcal{E}^-_s, s \in \mathcal{S} \label{xUpate}\\
    x_{v_iv_j}^*\overset{\vartriangle}{=}  x_{v_iv_j}+\max_{s \in \mathcal{S}} x^s_{v_iv_j} \quad \forall  (v_i,v_j) \in \mathcal{E}^+ \label{xUpate2}.
\end{align}
The right hand side of Eq. \eqref{xUpate2} cannot exceed $1$ at optimality because of the constraint in Eq.~\eqref{QPhiDef}.
%


Given the solution $x_{v_iv_j}^*$, the optimizing solution to each Benders subproblem $s$ is denoted $x_{v_iv_j}^{s*}$ and is defined as follows.
\begin{equation}
x_{v_iv_j}^{s*}=
\left\{
\begin{aligned}
1, & \quad \text{if~} (v_i,v_j) \in \color{black} \mathcal{E}^-_s\color{black} \\
0, & \quad \text{otherwise}.
\end{aligned}
\right.
\end{equation}
In Sec.~\ref{Sec_proof_Q_0}  in the supplement, we show that the cost of $\{x_{v_iv_j}^*, (x_{v_iv_j}^{s*})_{s \in \mathcal{S}}\}$ is no greater than that of $\{x_{v_iv_j} , (x_{v_iv_j}^s)_{s \in \mathcal{S}}\}$, with regard to the objective in Eq.~\eqref{objAug} and that $Q(\phi,s,\mathbf{x}^*)=0$ holds for all $s \in \mathcal{S}$. \\
It follows that there always exists an optimizing solution $\mathbf{x}$ to Eq.~\eqref{objAug} such that $Q(\phi,s,\mathbf{x})=0$ for all $s \in \mathcal{S}$.

\textcolor{black}{Observe, that there exists an optimal partition $\mathbf{x}^s$ of the nodes of the graph \color{black}, in Eq. \eqref{QPhiDef}, which is 2-colorable. This is because any partition $\mathbf{x}^s$ can be altered without increasing its cost, by merging connected components that are adjacent to one another, not including the root node $v_s$.  Note, that merging any pair of such components, does not increase the cost, since those components are not separated by negative weight edges in subproblem $s$ and so the result is still a partition.\color{black} }


Given this observation, we rewrite the optimization Eq.~ \eqref{objAug} regarding $Q(\phi,s,\mathbf{x})$, using the node labeling formulation of min-cut, with the notation below.

\textcolor{black}{We indicate }\textcolor{black}{with $m_v=1$ that} \textcolor{black}{ node $v \in \mathcal{V}$ is not in the component associated with the root of subproblem $s$ and $m_v=0$ otherwise. To avoid extra notation $m_{v_s}$ is replaced by $0$. Let 
\begin{equation}
f^s_{v_iv_j}=
\left\{
\begin{aligned}
1, & \quad \text{for~} (v_i,v_j) \in \color{black} \mathcal{E}^+\color{black}, \textnormal{if~} (v_i,v_j) \text{~is cut in~ } \mathbf{x}^s, \text{~but is not cut in~} \mathbf{x}  \\
1, & \quad  \text{for~} (v_i,v_j) \in \color{black} \mathcal{E}^-_s \color{black}, \textnormal{if~} (v_i,v_j) \text{~is not cut in~ } \mathbf{x}^s, \text{~but is cut in~} \mathbf{x}. \\
\end{aligned}
\right.
\end{equation}
Thus, the definition for the first/second case implies a penalty of $\phi_{v_iv_j}$/ - $\phi_{v_iv_j}$, which is added to $Q(\phi,s,\mathbf{x})$.  Note moreover that $x^s_{v_iv_j}=f^s_{v_iv_j}$ for all  $(v_i,v_j) \in \mathcal{E}^+$ and that  $x^s_{v_iv_j}=1-f^s_{v_iv_j}$ for all  $(v_i,v_j) \in \mathcal{E}^-_s$.}

%
%
    %
    %
    %
    %
    %
    %
 Below we write $Q(\phi,s,\mathbf{x})$ as primal/dual LP, with primal constraints associated with dual variables $\psi,\lambda$, which are noted in the primal. Given binary $\mathbf{x}$, we need only enforce that $f,m$ are non-negative to ensure that there exists an optimizing solution for $f,m$ which is binary. This is a consequence of the optimization being totally unimodular, given that $\mathbf{x}$ is binary. Total unimodularity is a known property of the min-cut/max flow LP \cite{ford1956maximal}.
 The primal subproblem is therefore given by the following.
\begin{eqnarray}
\label{primalSub}
Q(\phi,s,\mathbf{x}) &=& \min_{\substack{f^s_{v_iv_j}  \geq 0\\ m_v \geq 0}}\sum_{(v_i,v_j) \in \mathcal{E}^+}\phi_{v_iv_j}f^s_{v_iv_j}  
 -\sum_{(v_s,v) \in \mathcal{E}^-_s}\phi_{v_s v} f^s_{v_s v} \\ 
\lambda^-_{v_iv_j} &:& m_{v_i}-m_{v_j} \leq  x_{v_iv_j}+f^s_{v_iv_j} \;   \quad  \forall (v_i,v_j) \in (\mathcal{E}^+ \setminus \mathcal{E}^+_s), \nonumber \\
\lambda^+_{v_iv_j} &:&   m_{v_j}-m_{v_i} \leq x_{v_iv_j}+f^s_{v_iv_j}\;  \quad \forall (v_i,v_j) \in (\mathcal{E}^+ \setminus \mathcal{E}^+_s),  \nonumber \\
\psi^-_v &:& x_{v_s v}-f^s_{v_s v} \leq m_v  \quad  \forall (v_s,v) \in \mathcal{E}^-_{s}  , \nonumber  \\
 \psi^+_{v} &:& m_v \leq x_{v_s v}+f^s_{v_s v}  \quad  \forall (v_s, v) \in \mathcal{E}^+_{s}  ,\nonumber 
\end{eqnarray}
This yields to the corresponding dual subproblem.
\begin{align}
\label{BendersSub}
\max_{\substack{\lambda \geq 0\\ \psi \geq 0}} ~ -\sum_{(v_i,v_j) \in  (\mathcal{E}^+ \setminus \mathcal{E}^+_s)}(\lambda^-_{v_iv_j}+\lambda^+_{v_iv_j})x_{v_iv_j} 
 +\sum_{(v_s,v) \in \mathcal{E}^-_s}\psi^-_{v}x_{v_s v}-\sum_{(v_s,v) \in \mathcal{E}^+_s}\psi^+_{v}x_{v_s v}  
\end{align}
\begin{multline}
\mbox{s.t. } \quad \psi^+_{v_i}~ \mathbbm{1}_{\mathcal{E}^+_s}(v_s,v_i)-\psi^-_{v_i}~ \mathbbm{1}_{\mathcal{E}^-_s}(v_s,v_i) +\\ \sum_{\substack{v_j \\(v_i,v_j) \in  (\mathcal{E}^+ \setminus \mathcal{E}^+_s)
}}(\lambda^-_{v_iv_j}-\lambda^+_{v_iv_j})  
+ \sum_{\substack{v_j \\ (v_j,v_i) \in ( \mathcal{E}^+ \setminus \mathcal{E}^+_s)}} (\lambda^+_{v_jv_i}-\lambda^-_{v_jv_i})\geq 0\quad \quad \forall v_i \in \mathcal{V}-v_s \nonumber 
\end{multline}
\begin{eqnarray}
-\phi_{v_s v}-\psi^-_{v} &\geq& 0\quad \quad \forall  (v_s,v) \in \mathcal{E}^-_s \nonumber \\
\phi_{v_s v}-\psi^+_{v} &\geq& 0\quad \quad \forall (v_s,v) \in \mathcal{E}^+_s \nonumber \\
\textcolor{black}{\phi_{v_iv_j}-(\lambda^-_{v_iv_j} +\lambda^+_{v_iv_j} )} &\geq& 0\quad \quad \forall (v_i,v_j) \in \textcolor{black}{(\mathcal{E}^+ \setminus \mathcal{E}^+_s)}. \nonumber 
\end{eqnarray}

In Eq.~\eqref{BendersSub} and subsequently $\mathbbm{1}_{\Lambda}(x)$ denotes the binary indicator function for some set $\Lambda$, which returns one if $(x \in \Lambda)$ and zero otherwise. We now consider the constraint that $Q(\phi,s,\mathbf{x})=0$.  Note that any dual feasible solution for the dual problem~\eqref{BendersSub} describes an affine function of $\mathbf{x}$, which is a tight lower bound on $Q(\phi,s,\mathbf{x})$. We compact the terms $\lambda,\psi$ into $\omega^z$,  where $\omega^z_{v_iv_j}$ is associated with the $x_{v_iv_j}$ term.
\begin{equation*}
    \omega^z_{v_iv_j} =
    \left\{
    \begin{aligned}
    -(\lambda^-_{v_iv_j}+\lambda^+_{v_iv_j}) ,  &\quad \quad\text{if~} (v_i,v_j) \in (\mathcal{E}^+ \setminus \mathcal{E}^+_s) \\
    -\psi^+_{v_j}, & \quad \quad  \text{if~}  (v_i,v_j) \in \mathcal{E}^+_s  \\
    \psi^-_{v_j}, &\quad \quad   \text{if~} (v_i,v_j)\in \mathcal{E}^-_s \\
     0, &  \quad \quad \textcolor{black}{\text{if~} (v_i,v_j) \in (\mathcal{E}^- \setminus \mathcal{E}^-_s)} .  
    \end{aligned}
    \right.
 \end{equation*}
We denote the set of all dual feasible solutions across $s \in \mathcal{S}$ as $\mathcal{Z}$, with $z \in \mathcal{Z}$.  Observe, that to enforce that $Q(\phi,s,\mathbf{x})=0$, it is sufficient to require that $\sum_{(v_i,v_j) \in \mathcal{E}} x_{v_iv_j}\omega^z_{v_iv_j} \leq 0$, for all $ z \in \mathcal{Z}$. We formulate CC as optimization using $\mathcal{Z}$ below.  
\begin{align}
\label{masterBend0}
 \eqref{objAug} \rightsquigarrow  \eqref{masterBend0} &= 
\min_{\substack{\mathbf{x} \in \{0,1\}^{\vert \mathcal{E}\vert }}}\sum_{ (v_i,v_j) \in \mathcal{E}^+} \phi_{v_iv_j}x_{v_iv_j}-\sum_{(v_i, v_j) \in \mathcal{E}^-}(1-x_{v_iv_j})\phi_{v_iv_j}   \tag{$\mathrm{CC_3}$}\\
\mbox{ s.t. } & \sum_{(v_i,v_j) \in \mathcal{E}} x_{v_iv_j}\omega^z_{v_iv_j} \leq  0  \quad \quad \forall z \in \mathcal{Z} \nonumber
\end{align}
\subsection{Cutting Plane Optimization}
Optimization in Eq.~\eqref{masterBend0} is intractable since $|\mathcal{Z}|$ equals the number of dual feasible solutions across subproblems, which is infinite. Since we cannot consider the entire set $\mathcal{Z}$, we use a cutting plane approach to construct a set $\hat{\mathcal{Z}} \subset \mathcal{Z}$, that is sufficient to solve Eq.~\eqref{masterBend0} exactly. We initialize $\hat{\mathcal{Z}}$ as the empty set.  We iterate between solving the LP relaxation of Eq.~\eqref{masterBend0} over $\hat{\mathcal{Z}}$ (referred to as the master problem) and generating new Benders rows 
until no violated constraints exist.  

This ensures that no violated cycle inequalities exist but may not ensure that $\mathbf{x}$ is integral. To enforce integrality, we iterate between solving the ILP in Eq.~\eqref{masterBend0} over $\hat{\mathcal{Z}}$ and adding Benders rows to $\hat{\mathcal{Z}}$. By solving the LP relaxation first, we avoid unnecessary and expensive calls to the ILP solver. 

To generate Benders rows given $\mathbf{x}$, we iterate over $\mathcal{S}$ and generate one Benders row using Eq.~\eqref{BendersSub}, if $s$ is associated with a violated cycle inequality, 
which we determine as follows.  
Given $s,\mathbf{x}$ we iterate over $(v_i,v_j)\in \mathcal{E}^-_s$.  We find the shortest path from $v_i$ to $v_j$ on graph $\mathcal{G}$ with edges $\mathcal{E}$, with weights equal to the vector $\mathbf{x}$. If the length of this path, denoted as $d(v_i,v_j)$, is strictly less than $x_{v_iv_j}$, then we have identified a violated cycle inequality associated with $s$.

We describe our cutting plane approach in Alg.~\ref{Alg_BasicBend_4}, with line by line description in Sec.~\ref{Sec_lineByLine} in the supplementary material. To accelerate optimization, we add MWR in addition to standard Benders rows, which we describe in the following Sec.~\ref{Sec_pareto}.

Prior to termination of  Alg.~\ref{Alg_BasicBend_4}, one can produce a feasible integer solution $\mathbf{x}^*$ from any solution $\mathbf{x}$, provided by the master problem, as follows. First, for each $(v_i,v_j) \in \mathcal{E}$, set $x^{**}_{v_iv_j}=1$, if $x_{v_iv_j}>\frac{1}{2}$ and otherwise set $x^{**}_{v_iv_j}=0$.  Second, for each $(v_i,v_j) \in \mathcal{E}$, set $x^*_{v_iv_j}=1$, if  $v_i,v_j$ are in separate connected components of the solution described by $\mathbf{x}^{**}$ and otherwise set $x^*_{v_iv_j}=0$.  The cost of the feasible integer solution $\mathbf{x}^*$ provides an upper bound on the cost of the optimal solution. In Sec.~\ref{Sec_Anytime} (supplementary material), we provide a more involved approach to produce feasible integer solutions. 

In this section, we characterized CC using Benders decomposition and provided a cutting plane algorithm to solve the corresponding optimization. 
\begin{algorithm}[t]
\caption{Benders Decomposition for CC (BDCC)}
\begin{algorithmic}[1]
\State $\hat{\mathcal{Z}} = \{ \}$
\label{alg_4_init}
\State done\_LP = False
\label{alg_4_init_end}
\Repeat
\label{alg_4_loop}
\State $\mathbf{x} = $ Solve Eq.~\eqref{masterBend0} over $\hat{\mathcal{Z}}$ enforcing integrality if and only if  done\_LP=True
\label{alg_4_rmp_end}
\State did\_add = False
\label{alg_4_did_add}
\For{$s \in \mathcal{S}$}
\label{alg_4_pricing}
\If{ $\exists (v_i,v_j) \in \mathcal{E}^-_s $ s.t. $ d(v_i,v_j)<x_{v_iv_j}$}
\label{alg_4_do_check}
\State $z_1$ = Get Benders row via Eq \eqref{BendersSub}.
\label{alg_4_pr_1}
\State  $z_2$ = Get MWR via Sec. \ref{Sec_pareto}.
\label{alg_4_pareto}
%
\State $\hat{\mathcal{Z}} = \hat{\mathcal{Z}}\cup z_1 \cup z_2$ 
\label{alg_4_add}
\State did\_add = True
\label{alg_4_did_add_itt}
\EndIf
\EndFor
\label{alg_4_pricing_end}
\If{ did\_add=False }
\label{alg_4_swap_to_ILP}
\State done\_LP = True
\EndIf
\label{alg_4_swap_to_ILP_end}
 \Until{ did\_add=False AND $x_{v_iv_j} \in \{0,1\}  ~ \forall (v_i,v_j)\in \mathcal{E}$  }
\label{alg_4_loop_end}
\State Return $\mathbf{x}$
\label{alg_4_ret_x}
\end{algorithmic}
\label{Alg_BasicBend_4}
 \end{algorithm}
\section{Magnanti-Wong Benders Rows}
\label{Sec_pareto}
We accelerate Benders decomposition (see~ Sec. \ref{Sec_BendersForm}) using the classic operations research technique of Magnanti-Wong Benders Rows (MWR) \cite{magnanti1981accelerating}. The Benders row, given in Eq.~\eqref{BendersSub}, provides a tight bound at $\mathbf{x}^*$, where $\mathbf{x}^*$ is the master problem solution used to generate the Benders row. However, ideally, we want our Benders row to provide good lower bounds for a large set of $\mathbf{x}$ different from $\mathbf{x}^*$, while being tight (or perhaps very active) at $\mathbf{x}^*$. To achieve this, we use a modified version of Eq.~\eqref{BendersSub}, where we replace the objective and add one additional constraint.  

We follow the tradition of the operations research literature and use a random negative valued vector (with unit norm) in place of the objective Eq.~\eqref{BendersSub}. This random vector is unique each time a Benders subproblem is solved. We experimented with using as an objective $\frac{-1}{.0001+ |\phi_{v_iv_j}|}$, which encourages the cutting of edges with large positive weight, but it works as well as the random negative objective. Here $.0001$ is a tiny positive number. It prevents the terms in the objective from becoming infinite.

Below, we enforce the new Benders row to be active at $\mathbf{x}^*$, by requiring that the dual cost is within a tolerance $\tau \in (0,1)$ of the optimum w.r.t. the objective in Eq.~\eqref{BendersSub}.
\begin{align}
\label{new_feas_const}
\tau Q(\phi,s,\mathbf{x}) \leq~ -\sum_{(v_i,v_j) \in \textcolor{black}{ (\mathcal{E}^+ \setminus \mathcal{E}^+_s) }}(\lambda^-_{v_iv_j}+\lambda^+_{v_iv_j})x_{v_iv_j} 
+\sum_{(v_s,v) \in \mathcal{E}^-_s}\psi^-_{v}x_{v_s v}-\sum_{(v_s ,v) \in \mathcal{E}^+_s}\psi^+_{v}x_{v_s v} 
%
\end{align}
      Here, $\tau=1$ requires optimality w.r.t. the objective in Eq.~\eqref{BendersSub}, while $\tau=0$ ignores optimality. In our experiments, we found that $\tau=\frac{1}{2}$ provides strong performance.


\begin{figure}[t]
  \centering
  \begin{tabular}{@{\hspace{-0.5cm}}c@{\hspace{-0.5cm}}c@{}}
    \includegraphics[width=0.55\textwidth]{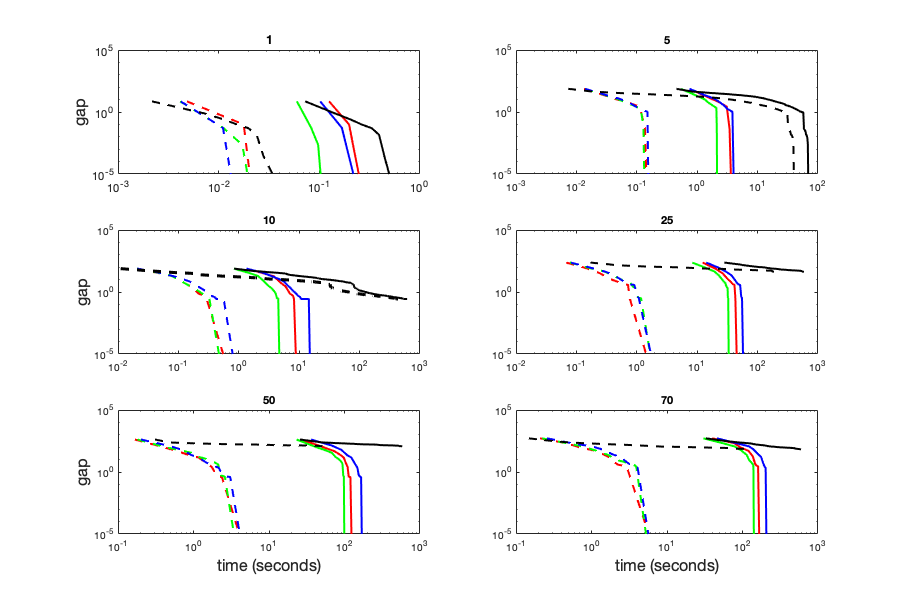}&
    \includegraphics[width=0.55\textwidth]{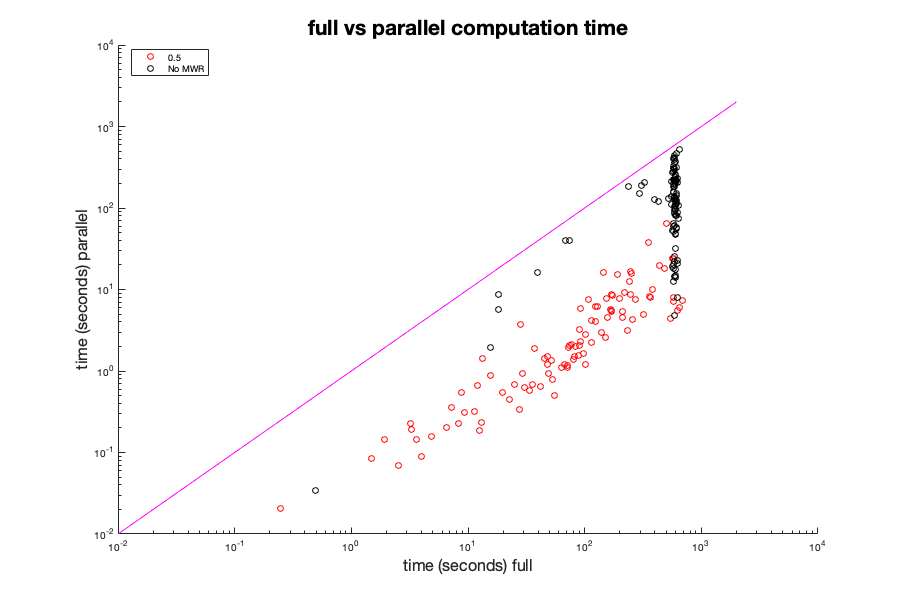}
    \end{tabular}
    \caption{\textbf{Left:}  We plot the gap between the upper and lower bounds as a function of time for various values of $\tau$ on selected problem instances.  We use red,green,blue for $\tau=[0.5,0.99,.01]$ respectively, and black for not using Magnanti-Wong rows.  We show both the computation time with and without exploiting parallelization of subproblems with dotted and solid lines, respectively.  We use titles to indicate the approximate difficulty of the problem as ranked by input file size of 100 files.  \newline
    \textbf{Right:} We compare the benefits of parallelization and MWR across our data set. We scatter plot the total running time versus the total running time when solving each subproblem is done on its own CPU across problem instances. We use red to indicate $\tau=0.5$ and black to indicate that MWR are not used. We draw a line with slope=1 in magenta to better enable appreciation of the red and black points. 
    NOTE:  The time spent generating Benders rows, in a given iteration of BDCC when using parallel processing, is the maximum time spent to solve any sub-problem for that iteration.} 
    \label{fig:my_plots}
\end{figure}
\section{Experiments: Image Segmentation}
\label{Sec_Exper}
In this section, we demonstrate the value of our algorithm BDCC on CC problem instances for image segmentation on the benchmark Berkeley Segmentation Data Set (BSDS) \cite{bsdspaper}.  Our experiments demonstrate the following three findings.  (1)  BDCC solves CC instances for image segmentation; (2) BDCC successfully exploits parallelization; (3)  the use of MWR dramatically accelerates optimization.  

To benchmark performance, we employ cost terms provided by the OPENGM2 dataset~\cite{andres2014opengm2} for BSDS. This allows for a direct comparison of our results to the ones from \citet{ilpalg}.
We use the random unit norm negative valued objective when generating MWR.  We use CPLEX to solve all linear and integer linear programming problems considered during the course of optimization.   
We use a maximum total CPU time of 600 seconds, for each problem instance (regardless of parallelization).   

We formulate the selection of $\mathcal{S}$, as a minimum vertex cover problem, where for every edge $(v_i,v_j) \in \mathcal{E}^-$, at least one of $v_i,v_j$ is in $\mathcal{S}$.  We solve for the minimum vertex cover exactly as an ILP. Given $\mathcal{S}$, we assign edges in $\mathcal{E}^-$ to a connected selected node in $\mathcal{S}$ arbitrarily. We found experimentally that solving for the minimum vertex cover consumed negligible CPU time for our dataset. We attribute this fact to the structure of our problem domain, since the minimum vertex cover is an NP-hard problem. For problem instances where solving for the minimum vertex cover exactly is difficult, the minimum vertex cover problem can be solved approximately or greedily.

In Fig.~\ref{fig:my_plots} (left) we demonstrate the effectiveness of BDCC with various $\tau$ for different problem difficulties.  We observe that the presence of MWR dramatically accelerates optimization. However, the exact value of $\tau$ does not effect the speed of optimization dramatically. We show performance with and without relying on parallel processing.  Our parallel processing times assume that we have one CPU for each subproblem.  For the problem instances in our application the number of subproblems is under one thousand, each of which are very easy to solve. The parallel and non-parallel time comparisons share only the time to solve the master problem. We observe large benefits of parallelization for all settings of $\tau$. However, when MWR are not used, we observe diminished improvement, since the master problem consumes a larger proportion of total CPU time. 

In Fig.~\ref{fig:my_plots}(right), we demonstrate the speed up induced by the use of parallelization. 
For most problem instances, the total CPU time required when using no MWR was prohibitively large, which is not the case when MWR are employed.  Thus most problem instances solved without MWR terminated early. 

In Tab.~\ref{tab:my_label}, we consider the convergence of the bounds for $\tau=\{0,\frac{1}{2} \}$; ( $\tau=0$ means that no MWR are generated). We consider a set of tolerances on convergence regarding the duality gap, which is the difference between the anytime solution (upper bound) and the lower bound on the objective. For each such tolerance $\epsilon$, we compute the percentage of instances, for which the duality gap is less than $\epsilon$, after various amounts of time.  We observe that the performance of optimization without MWR, but exploiting parallelization performs worse than using MWR, but without paralleliziation.  This demonstrates that, across the dataset, MWR are of greater importance than parallelization.  

\begin{table}
  \caption{We show the percentage of problems solved that have a duality gap of up to tolerance $\epsilon$,  within a certain amount of time (10,50,100,300) seconds, with and without MWR/parallelization.  We use par =1 to indicate the use of parallelization and par=0 otherwise.  Here $\tau=0$ means that no MWR are generated.}
  \label{tab:my_label}
  \centering
  \small
 \begin{tabular}{c c c |c c c c}
 
    \toprule
   $\epsilon$=0.1&$\tau$ & par & 10 & 50 & 100 & 300\\
        \midrule
&0.5&0&0.149&0.372&0.585&0.894\\
&0&0&0.0106&0.0532&0.0745&0.106\\
&0.5&1&0.266&0.777&0.904&0.968\\
&0&1&0.0426&0.0745&0.0745&0.138\\
\midrule
\midrule
$\epsilon$=1&$\tau$ & par & 10 & 50 & 100 & 300\\
\midrule
&0.5&0&0.149&0.394&0.606&0.904\\
&0&0&0.0106&0.0638&0.0745&0.16\\
&0.5&1&0.319&0.819&0.947&0.979\\
&0&1&0.0532&0.0745&0.106&0.17\\
\midrule
\midrule
$\epsilon$=10& $\tau$ & par & 10 & 50 & 100 & 300\\
\midrule
&0.5&0&0.202&0.426&0.628&0.915\\
&0&0&0.0532&0.0957&0.128&0.223\\
&0.5&1&0.447&0.936&0.979&0.989\\
&0&1&0.0638&0.128&0.181&0.287\\
    \bottomrule
  \end{tabular}
\end{table}

\section{Conclusions}
\label{Sec_conc}

We present a novel methodology for finding optimal correlation clustering in arbitrary graphs. Our method exploits the Benders decomposition to avoid the enumeration of a large number of cycle inequalities. This offers a new technique in the toolkit of linear programming relaxations, that we expect will find further use in the application of combinatorial optimization to problems in computer vision. 

The exploitation of results from the domain of operations research may lead to improved variants of BDCC.  For example, one can intelligently select the subproblems to solve instead of solving all subproblems in each iteration. This strategy is referred to as partial pricing in the operations research literature.  Similarly one can devote a minimum amount of time in each iteration to solve the master problem so as to enforce integrality on a subset of the variables of the master problem.  

%
\small

\bibliography{example_paper,papers}
\bibliographystyle{abbrvnat}
 
 \newpage

\appendix

\section{APPENDIX:  $Q(\phi,s,\mathbf{x}^*)=0$ at Optimality}
\label{Sec_proof_Q_0}
%

In this section, we demonstrate that there exists an $\mathbf{x}^*$, that minimizes Eq.~\eqref{objAug}, for which $Q(\phi,s,\mathbf{x}^*)=0$. Given an arbitrary solution $\{x_{v_iv_j} , (x_{v_iv_j}^s)_{s \in \mathcal{S}}\}$ another solution $\{x^*_{v_iv_j} , (x_{v_iv_j}^{s*})_{s \in \mathcal{S}}\}$ is constructed, for which $Q(\phi, s, \mathbf{x}^*)=0$ holds, without increasing the objective in Eq.~\eqref{objAug}. We write the updates below in terms of $\mathbf{x}^s$. 
\begin{equation}
\label{changeEq}
\begin{split} 
x_{v_iv_j}^* &~\overset{\vartriangle}{=}  x_{v_iv_j}+\max_{\textcolor{black}{s \in \mathcal{S} }}x^s_{v_iv_j} \quad \forall  (v_i,v_j) \in \mathcal{E}^+\\
x_{v_iv_j}^*& ~\overset{\vartriangle}{=}  ~x_{v_iv_j}+x^s_{v_iv_j}-1 \quad \forall (v_i,v_j) \in \mathcal{E}^-_s, s \in \mathcal{S} \\
x^{s*}_{v_iv_j}& ~\overset{\vartriangle}{=} ~0 \quad \forall  (v_i,v_j) \in \mathcal{E}^+ \\
x^{s*}_{v_iv_j}& ~ \overset{\vartriangle}{=} ~1 \quad \forall  (v_i,v_j) \in \mathcal{E}^-_s, s \in \mathcal{S}.
\end{split}
\end{equation}
The updates in Eq. \eqref{changeEq} are equivalent to the following updates using $f^s$,$f^{s*}$. Here $f^s,f^{s*}$ correspond to the optimizing solution for $f$ in subproblem $s$, given $\mathbf{x},\mathbf{x}^*$ respectively.
\begin{equation}
\label{changeEqF}
\begin{split}
 x^*_{v_iv_j} &= x_{v_iv_j} +\max_{\textcolor{black}{s \in \mathcal{S}}}  f^s_{v_iv_j} \quad \forall (v_i,v_j) \in \mathcal{E}^+ \\
 x^*_{v_iv_j} &=  x_{v_iv_j}-f^s_{v_iv_j} \quad \forall  (v_i,v_j) \in \mathcal{E}^-_s, s \in \mathcal{S}   \\
 f^{s*}_{v_iv_j}&= 0 \quad \forall  (v_i,v_j) \in \mathcal{E}^+ \\
 f^{s*}_{v_iv_j}&= 0 \quad \forall  (v_i,v_j) \in \mathcal{E}^-_s 
\end{split}
\end{equation}

These updates in Eq. \eqref{changeEq} and Eq. \eqref{changeEqF} preserve the feasibility of the primal LP in Eq. \eqref{primalSub}.  Also notice, that since $f^{s*}$ is a zero valued vector for all $s\in \mathcal{S}$, then  $Q(\phi,s,x^*)=0$ for all $s \in \mathcal{S}$.  

We now consider, the total change in Eq. \eqref{objAug} corresponding to edge $(v_i,v_j)\in \mathcal{E}^+$, induced by Eq. \eqref{changeEq}, which is non-positive. 
The objective of the master problem increases by $\phi_{v_iv_j}\max_{s \in \mathcal{S}}x^s_{v_iv_j}$, while the total decrease in the objectives of the subproblems is $\phi_{v_iv_j}\sum_{s\in \mathcal{S}} x^s_{v_iv_j}$.  Since the latter value is greater than the former value, the total change in problem \eqref{objAug} decreases more than it increases. 
Considering on the other hand the total change of Eq.~ \eqref{objAug} corresponding to edge $(v_i,v_j)\in \mathcal{E}^-$, induced by Eq. \eqref{changeEq}, which is zero, yields in an increase  
of the objective of the master problem by $-\phi_{v_iv_j}(1-x^n_{v_iv_j})$, while the objective of subproblem $s$ decreases by $-\phi_{v_iv_j}(1-x^s_{v_iv_j})$.  
This shows that the objective of Eq. \eqref{objAug} is minimized for $\mathbf{x}^*$.
\section{Line by Line Description of BDCC}
\label{Sec_lineByLine}
We provide the line by line description of Alg. \ref{Alg_BasicBend_4}. 
  \begin{itemize}
      \item Line \ref{alg_4_init}:  Initialize the nascent set of Benders rows $\hat{\mathcal{Z}}$ to the empty set.  
      \item Line \ref{alg_4_init_end}:  Indicate that we have not solved the LP relaxation yet.  
      \item Line \ref{alg_4_loop}-\ref{alg_4_loop_end}: Alternate between solving the master problem and generating Benders rows, until a feasible integral solution is produced.
      \begin{enumerate}
          \item Line \ref{alg_4_rmp_end}: Solve the master problem providing a solution $\mathbf{x}$, which may not satisfy all cycle inequalities.  We enforce integrality if we have finished solving the LP relaxation, which is indicated by done\_lp=True.  
          \item Line \ref{alg_4_did_add}:  Indicate that we have not yet added any Benders rows to this iteration.
          \item Line \ref{alg_4_pricing}-\ref{alg_4_pricing_end}:  Add Benders rows by iterating over  subproblems and adding Benders rows corresponding to subproblems, associated with violated cycle inequalities.
          \begin{itemize}
          \item Line \ref{alg_4_do_check}:  Check if there exists a violated cycle inequality associated with $\mathcal{E}^-_s$.  This is done by iterating over $(v_i,v_j) \in \mathcal{E}^-_s$ and checking if the shortest path from $v_i$ to $v_j$ is less than $x_{v_iv_j}$.  This distance is defined on the graph's edges $\mathcal{E}$ with weights equal to $\mathbf{x}$.
          \item Lines \ref{alg_4_pr_1}-\ref{alg_4_add}:  Generate Benders rows associated with subproblem $s$ and add them to nascent set $\hat{\mathcal{Z}}$.  
          \item Line \ref{alg_4_did_add_itt}:  Indicate that a Benders row was added this iteration.  
          \end{itemize}
          \item Lines \ref{alg_4_swap_to_ILP}-\ref{alg_4_swap_to_ILP_end}:  If no Benders rows were added to this iteration, we enforce integrality on $\mathbf{x}$, when solving the master problem for the remainder of the algorithm.
      \end{enumerate}
      \item Line \ref{alg_4_ret_x} Return solution $\mathbf{x}$.
  \end{itemize}
\section{Generating Feasible Integer Solutions Prior to Convergence}
\label{Sec_Anytime}

Prior to the termination of optimization, it is valuable to provide feasible integer solutions on demand.  This is so that a practitioner can terminate optimization, when the gap between the objectives of the integral solution and the relaxation is small.  In this section we consider the production of feasible integer solutions, given the current solution $\mathbf{x}^*$ to the master problem, which may neither obey cycle inequalities or be integral.  We refer to this procedure as rounding.

Rounding is a coordinate descent approach defined on the graph \textcolor{black}{$\mathcal{G}$ and its edges $\mathcal{E}$} with weights $\kappa$, determined using $\mathbf{x}^*$ below. 
\begin{align}
\label{kappaEq}
     \kappa_{v_iv_j}&=\phi_{v_iv_j}(1-x^*_{v_iv_j})\quad \forall (v_i,v_j) \in \mathcal{E}^+\\
      \kappa_{v_iv_j}&=\phi_{v_iv_j} x^*_{v_iv_j} \quad \forall (v_i,v_j) \in \mathcal{E}^-\nonumber
\end{align}
%
  Consider that $\mathbf{x}^*$ is integral and feasible (where feasibility indicates that $\mathbf{x}^*$ satisfies all cycle inequalities).  Let $\mathbf{x}^{s*}$ define the boundaries in partition $\mathbf{x}^*$, of the connected component containing $s$. Here $x^{s*}_{v_iv_j}=1$ if exactly one of $v_i,v_j$ is in the connected component containing $s$ under cut $\mathbf{x}^*$. Observe, that $Q(\kappa,s,\mathbf{x}^{0s})=0$, where $x^{0s}_{v_iv_j}=\mathbbm{1}_{\mathcal{E}^-_s}(v_i,v_j)$, is achieved using $\mathbf{x}^{s*}$ as the solution to Eq. \eqref{primalSub}.  Thus $\mathbf{x}^{s*}$ is the minimizer of Eq. \eqref{primalSub}.  The union of the edges cut in $\mathbf{x}^{s*}$ across $s \in \mathcal{S}$ is identical to $\mathbf{x}^*$.  Note that when $\mathbf{x}^*$ is integral and feasible then the solution produced below has cost equal to that of $\mathbf{x}^*$.  
  \begin{align}
  \label{ezProcR}
      \mathbf{x}^{s*} & ~\overset{\vartriangle}{=} \mbox{ minimizer of } Q(\kappa,s,\mathbf{x}^{0s}) \quad  \forall s \in \mathcal{S}\nonumber\\
      x^+_{v_iv_j} & ~\overset{\vartriangle}{=} \max_{s \in \mathcal{S}}x^{s*}_{v_iv_j} \quad \forall (v_i,v_j) \in \mathcal{E}^+  \\
      x^+_{v_iv_j}&  ~\overset{\vartriangle}{=} x^{s*}_{v_iv_j} \quad \forall (v_i,v_j) \in \mathcal{E}^-_s, s \in \mathcal{S} \nonumber 
  \end{align}
  %
  %
  %
  %
 The procedure of Eq. \eqref{ezProcR} can be used regardless of whether $\mathbf{x}^*$ is integral or feasible.  Note that if $\mathbf{x}^*$ is close to integral and close to feasible, then Eq. \eqref{ezProcR} is biased to produce a solution that is similar to $\mathbf{x}^*$ by design of $\kappa$.

%
We now consider a serial version of Eq. \eqref{ezProcR}, which may provide improved results.  We construct a partition $\mathbf{x}^+$ by iterating over $s \in \mathcal{S}$, producing component partitions as in Eq. \eqref{ezProcR}. We alter $\kappa$ by allowing for the cutting of edges previously cut with cost zero.  We formally describe this serial rounding procedure below in Alg. \ref{Alg_UB}.  
%
%
%
%
\begin{algorithm}
 \caption{Generating an Integral and Feasible Solution Given Infeasible and or Non-Integral Input $\mathbf{x}^*$)}
\begin{algorithmic}[1]
\State  $x^+_{v_iv_j} = 0 \quad \forall (v_i,v_j) \in \mathcal{E}$
\label{alg_2_init_x}
\State $\kappa_{v_iv_j} = \phi_{v_iv_j}x^*_{v_iv_j}\quad \forall (v_i,v_j) \in \mathcal{E}^-$
\label{alg_2_init_eta}
\State $\kappa_{v_iv_j} = \phi_{v_iv_j}(1-x^*_{v_iv_j}) \quad \forall (v_i,v_j) \in \mathcal{E}^+$
\label{alg_2_init_eta_end}
\For{$s \in \mathcal{S}$}
\label{alg_2_loop}
\State $\mathbf{x}^s =$ minimizer for $Q(\kappa,s,\mathbf{x}^{0s})$ given fixed $\kappa,s$.
\label{alg_2_grab_x_s}
\State $x^+_{v_iv_j} = \max(x^+_{v_iv_j},x^s_{v_iv_j})\quad \forall (v_i,v_j) \in \mathcal{E}$
\label{alg_2_aug_x_star_neg}
\State $\kappa_{v_iv_j}= \kappa_{v_iv_j}(1-x^+_{v_iv_j}) \quad \forall (v_i,v_j) \in \mathcal{E}$
\label{alg_2_aug_eta}
\EndFor
\label{alg_2_loop_end}
\State Return $\mathbf{x}^+$
\label{alg_2_ret_x}
\end{algorithmic}
\label{Alg_UB}
  \end{algorithm}
\begin{itemize}
    \item Line \ref{alg_2_init_x}:  Initialize $\mathbf{x}^+$ as the zero vector.  
    \item Line \ref{alg_2_init_eta}-\ref{alg_2_init_eta_end}: Set $\kappa$ according to Eq. \eqref{kappaEq}
    \item Line \ref{alg_2_loop}-\ref{alg_2_loop_end}:  Iterate over $s \in \mathcal{S}$ to construct $\mathbf{x}^+$ by cutting edges cut in the subproblem.
    \begin{enumerate}
        \item Line \ref{alg_2_grab_x_s}:  Produce the lowest cost cut  $\mathbf{x}^s$ given altered edge weights $\kappa$ for subproblem $s$. 
        \item Line \ref{alg_2_aug_x_star_neg}:  Cut edges in $\mathbf{x}^+$ that are cut in $\mathbf{x}^s$.
        \item Line \ref{alg_2_aug_eta}:  Set $\phi_{v_iv_j}$ to zero for cut edges in $\mathbf{x}^+$.
    \end{enumerate}
    \item Line \ref{alg_2_ret_x}:  Return the solution $\mathbf{x}^+$
\end{itemize}
 When solving for the fast minimizer of $Q(\kappa,s,\mathbf{x}^{0n})$, we rely on the network flow solver of \citet{QPBO}, though we do not exploit its capacity to tackle non-submodular problems.

\end{document}